\pdfoutput=1
\documentclass[11pt,a4paper]{article}

\usepackage[review]{EMNLP2023}

\usepackage{fixltx2e}
\usepackage{float}
\usepackage{booktabs}
\usepackage{amssymb}
\usepackage{multirow}
\usepackage{subfig}
\usepackage{caption}
\usepackage{graphicx}
 \usepackage{color}
 \usepackage{comment}
\usepackage{algorithm}
\usepackage{algpseudocode}
\usepackage{pgfplots}

\usepgfplotslibrary{external}
\usepackage{mathtools}
\usepackage{breqn}
\usepackage[T1]{fontenc}
\usepackage[utf8]{inputenc}

\usepackage{microtype}
\usepackage{multirow}
\usepackage{inconsolata}
\usepackage{booktabs,caption}
\usepackage[flushleft]{threeparttable}

\title{SNOiC: Soft Labeling and Noisy Mixup based Open Intent Classification Model}

\author{ Aditi Kanwar,  Aditi Seetha,  Satyendra Singh Chouhan \\
         MNIT Jaipur,  302017, INDIA \\ {\{2021pcp5102,2021rcp9548,sschouhan.cse\}@mnit.ac.in} \AND Rajdeep Niyogi  \\ IIT Roorkee, 247667, INDIA \\  rajdeep.niyogi@cs.iitr.ac.in } 
\begin{document}
\nolinenumbers
% \maketitle
{\makeatletter\acl@finalcopytrue
  \maketitle
}

\begin{abstract}
This paper presents a Soft Labeling and Noisy Mixup-based open intent classification model (SNOiC). Most of the previous works have used threshold-based methods to identify open intents, which are prone to overfitting and may produce biased predictions. 
Additionally, the need for more available data for an open intent class presents another limitation for these existing models. SNOiC combines Soft Labeling and Noisy Mixup strategies to reduce the biasing and generate pseudo-data for open intent class. The experimental results on four benchmark datasets show that the SNOiC model achieves a minimum and maximum performance of 68.72\% and 94.71\%, respectively, in identifying open intents. Moreover, compared to state-of-the-art models, the SNOiC model improves the performance of identifying open intents by  0.93\% (minimum) and 12.76\% (maximum). The model's efficacy is further established by analyzing various parameters used in the proposed model. An ablation study is also conducted, which involves creating three model variants to validate the effectiveness of the SNOiC model.  \\
\end{abstract}
%%----------------------------------------------------------------------------%%
\section{Introduction}
% Traditional Machine Learning , specially supervised learning trains model under the assumption of closed world . Under closed world  ,
% the model is trained based on assumption that classes seen during training ,  similar classes will be seen during testing . But in real world , model faces data from both seen and unseen classes . To overcome this limitation ,
% Open World Machine Learning came into picture. Under Open World Machine Learning , models are trained to identify both seen and unseen classes correctly. Open Intent Classification is one of the task solved using Open World Machine Learning Techniques.\\

% Open Intent Classification in a task-specific text based conversational system is a very challenging and critical task. In figure \ref{} , dialogue based system for Airlines is shown which covers 2 user intents , time and fare related queries. As can be observed , known i.e seen intents are classified correctly and intents other than that are classified as open intents. The task of open intent classification can generally be visualized as K+1 class classification problem where there are K known intent classes and one open intent class. The main challenge while training model is that data consist of only known intent classes.\\

Open intent classification is a Natural language processing (NLP) task where the objective is to train a model that can accurately identify the intent of a user's input and respond appropriately, such as routing the user's query to the appropriate service or providing a relevant response~\cite{parmar2023open}.
% that identifies the intention behind a user's input in an open domain setting. 
The term ``open" refers to the ability of the machine learning model to handle and respond to inputs that are outside of its training data or expected scenarios, i.e., it has not seen that specific intention before. 
% This is beneficial in situations with a large number of potential intents or where new intents may be introduced over time.

% Open intent Classification is a natural language processing (NLP) task identifying the intent behind a user's input in an open domain setting. \textbf{The term ``open"  typically refers to the idea that the model is able to handle and respond to inputs that are outside of its training data or expected scenarios.} In open intent classification, the system is trained to identify the intention of the user's input, even if it has not seen that specific intention before. This is useful in situations where there are a large number of potential intents or where new intents may be introduced over time.

% Open intent classification can be approached using machine learning techniques such as neural networks, support vector machines (SVMs), or decision trees. 

Machine learning methods, such as  neural networks (NN), decision trees (DT), and support vector machines (SVMs) can be employed to address open intent classification. In general, open intent classification can be thought of as a classification task with $M+1$ categories, including $M$ known intent categories and one category for open intents. The primary difficulty during model training is that the dataset only includes examples from known intent categories. 

In this line of research, the researchers focused on adjusting the decision boundary for each known intent class to detect outliers~\cite{chen2023empirical,bayer2022survey,cao2022survey,zeng2021adversarial,zhang2021deep,yan2020unknown}. One possible solution is to set a threshold on the prediction probability of the $M$ class classifier to determine whether a sample belongs to all known intents~\cite{shu2017doc}. Consequently, some researchers have utilized outlier detection techniques that are more adaptable to calculate and refine the boundary for making decisions. For example, in~\cite{yan2020unknown,lin2019deep}, the authors proposed learning deep discriminative features with the help of Gaussian mixture loss and margin cosine loss and then detect outliers, i.e., open intents using local outlier factor algorithms. Similarly, some authors presented a technique for acquiring distinct and informative deep features by applying self-supervised contrastive loss and large-margin cosine loss and then employing Mahalanobis distance to identify open intents \cite{xu2020deep,zeng2021adversarial}. In \cite{zhang2021deep}, the authors introduced a method that optimizes the decision boundary for outlier detection and feature learning jointly and adaptively.

In the works mentioned above, some deep learning models use a threshold value to detect open intent instances. However, they are prone to overfitting, and the $M$-class classifier may produce biased predictions, making it difficult to determine the threshold. Furthermore, some models only learn and tighten decision boundaries through available data for known intent classes. The absence of data for the open intent class also poses a limitation.

In order to mitigate the above limitations, we proposed SNOiC, an $M + 1$ open intent classification model. The model works in two steps. First, it is pre-trained using known intents, and second, it is trained for open intents using two strategies, Soft Labeling \cite{verma2019manifold} and Noisy Mixup \cite{lim2021noisy}. Soft Labeling reduces the biased predictions of the model for known intent classes by relocating probability from the known intent class to the open intent class. The Noisy Mixup  overcomes the impact of the unavailability of open intent data by generating pseudo-intent samples for the open intent class.
Noisy Mixup combines Manifold Mixup
% \cite{verma2019manifold} \cite{zhang2017mixup} 
and Noise Injection and has been used only for image and existing benchmark datasets. To the best of our knowledge, it has never been applied to the text dataset (intent classification). 
The main contributions of the paper can be summarized as follows.
\begin{enumerate}
    \item An $M+1$ open intent classification model, SNOiC, is proposed that comprises Soft Labeling and Noisy Mixup methods.
    % \item To the best of our knowledge, this is the first research in applying the Noisy Mixup for the textual dataset for the task of open intent classification.  
    \item The SNOiC model is comprehensively evaluated using four benchmark datasets and validated by comparing the performance with other state-of-the-art models. 
    \item An ablation study is performed to evaluate the effectiveness of the SNOiC model, which involved creating three model variants. 
    \item Model analysis is conducted to study the effects of various parameters in our proposed SNOiC model.
\end{enumerate}

\section{SNOiC: Open intent classification model}
This section presents the technical details of the proposed SNOiC model. 
% The model works in two steps. First, it is pre-trained using known intents (Section 2.1) and then trained for open intents using two strategies, Soft Labeling and Noisy Mixup (Section 2.2). 
The model undergoes a two-step training process. Section 2.1 describes the first step, which involves pre-training the model using known intents. Section 2.2, on the other hand, covers the second step, which involves training the model for open intent identification. 
% Formally, open intent classification task can be defined as follows. 

% \begin{definition}
%     Let  D = \{$(s_1,\ y_1),(s_2,\ y_2), \dots, (s_n,\ y_n )$\} represents the labeled intent training data for $M$ intent classes, where $s_i \in S$ is the $i^{th}$ example and $y_i \in $ \{$I_1$, \dots, $I_M$\} is class label (Intent). The open intent classification task is to categorize each test instance `$s$' as one of the $M$ classes or identifies it as an open class.
% \end{definition}

 % In this division, we provide the technical details of the proposed model. 
 % In this division, we first present the open intent classification task description. Then we present an in-depth description of our proposed work.
 SNOiC applies sentence transformer to 
extract features corresponding to the user query ($s_{i}$) of different intents. The sentence transformer functions in two steps. The \textit{Bidirectional Encoder Representations from Transformers} (BERT) \cite{kenton2019bert} model is used in the first stage to build contextualized word embedding for each user query. The word embedding is subjected to the mean pooling operation by BERT in the second stage to produce the sentence embedding. 
% Formally, given the \(i^{th}\) user query  \(s_i = \{t_1, t_2,\ldots,t_{l_i }\} \), which are applied to BERT model to compute the all the token embeddings \([CLS, T_1,\ldots,T_{l_i} ] \in R^{(l_i+1)\times H } \) from the last hidden layer of BERT, where CLS is the special classification token, \(l_i\) is the sentence length of the \(i^{th}\) user query, and $H$ is the hidden layer size [ ]. 
Formally, given user query \(s_i = \{t_1, t_2,\ldots,t_{l_i }\} \), where BERT model is utilized to compute the token embeddings \([CLS, T_1,\ldots,T_{l_i} ] \in R^{(l_i+1)\times H } \) from the final hidden layer of BERT. Here  $CLS$ denotes the special classification token, \(l_i\) represents the length of the \(i^{th}\) user query, and $H$ refers to the hidden layer size. The token embeddings produced from the final layer are transmitted to the subsequent layer, the mean pooling layer, to derive the average representation \(x_i \in R^H \), i.e.,
\begin{equation}
   x_i = mean\_pooling([CLS, T_1,\ldots,T_{l_i} ]) 
\end{equation}
% Next, \( x_i\) is passed to a dense layer $d$ to get the intent representation
The value of \( x_i\) is used as input to a dense layer $d$, which computes the intent representation
$e_i \in R^{dim}$ i.e.,
\begin{equation}
   e_i = ReLU(W_dx_i + b_d) 
\end{equation}

where $dim$ denotes the size of the intent representation, $W_d \in
R^{H\times dim }$ and  \(b_d\in R^{dim} \) represent the weight and bias  of layer $d$
respectively. This layer helps to further strengthen capability of feature extraction. 
% Next, the intent representation $e_{i}$ is used for pre-training using known intents.
% \subsection{Task Description}
% The aim of Open Intent Classification is to train model in such a way that is is able to :-
% \begin{itemize}
%     \item predict utterances of known intents correctly .
%     \item as well as predict utterances of open intents correctly .
% \end{itemize}

\subsection{Pre-training using known intents}
%  In this step, BERT extract features corresponding to user queries of different intents. Given the \(i^{th}\) user query  \(s_i = \{t_1, t_2,\ldots,t_{l_i }\} \), we can get
% all token embeddings \([CLS, T_1,\ldots,T_{l_i} ] \in R^{(l_i+1)\times H } \) from
% the last hidden layer of BERT, where CLS is the special classification token, \(l_i\) is the sentence length of the
% \(i^{th}\)
% user query, and $H$ is the hidden layer size [ ]. The token embeddings generated from the last layer are passed to the next layer (mean pooling layer)  to get the average representation \(x^i \in R^H \), i.e.,
% \begin{equation}
%    x_i = mean\_pooling([CLS, T_1,\ldots,T_{l_i} ]) 
% \end{equation}

% Next, \( x_i\) is passed to a dense layer $d$ to get the intent representation
% \(z_i \in R^D:\) ,i.e.,
% \begin{equation}
%    z_i = d(x_i) = ReLU(W_dx_i + b_d) 
% \end{equation}

% where $D$ is the dimension of the intent representation, \(W_d \in
% R^H\times D \) and  \(b_d\in R^D \) denote the weight and bias term of layer d
% respectively. This layer helps to further strengthen capability of feature extraction.

% \subsubsection{Pre-training using known intents}.
In this step, the model is trained using the labeled data of known intents from the training set. The intuition behind this is to learn and get a better intent representation $(e_{i})$ for the classification task. Softmax loss function \(L_s\) is used for learning intent representation.
\begin{equation}
\label{egn:softmaxloss}
     L_s  = -\frac{1}{N} \sum_{i=1}^{N}log{\frac{e^{\phi_M(e_i)^{y_i} }}{\sum_{j=1}^{M}e^{\phi_M(e_i)^{j} }}}
\end{equation}

% where S denotes the dimension of training data, \( \phi_M(·)\) is a $M$-class classifier, and \( \phi_M(·)^j\)  denote the j$th$ class   output logits . The classifier \( \phi_M(·)\) is a subset of \( \phi_{M+1}(·)\) and  corresponds to  the classifier for known intents classes.
where $N$ represents the size of the training data, and $\phi_M(.)$
denote an $M$-class classifier. The output logits of the $j^{th}$ class are represented by $\phi_M(·)^j$. 
The classifier $\phi_M(·)$ is a subset of $\phi_{M+1}(·)$ and pertains to the classifier for known intent classes.

\subsection{Training for open intent identification}
Training the model to identify open intent is challenging, as it requires more training data corresponding to open intent. Thus, one way to address this challenge is by generating pseudo-data for further model training. For this, we have utilized two strategies: Soft Labeling and Noisy Mixup.
% The former reduces overconfident predictions by modifying the label distribution of instances in the training set for known intents. The latter creates pseudo-data of an open intent class by combining noise generation and manifold mixup \cite{verma2019manifold},  \cite{zhang2017mixup}.
The technical details of both strategies are described below.

% Training the model to identify open intents is challenging as more training data corresponding to open intent is required. Thus, one way to handle the challenge is by generating pseudo-data for further model training. Therefore,
% for training we have used two strategies, Soft Labeling and Noisy Mixup. The former creates reduce overconfident
% prediction by models for known intents by modifying the label distribution of instances in training set for known intents. The latter creates pseudo-data of an open intent
% class by combining noise generation and manifold mixup  \cite{verma2019manifold} \cite{zhang2017mixup} . The technical details of both strategies are described below.
% For the model to detect and correctly classify  open intent as open intent, data specific to open intents is required; 
% One way is to generate pseudo data and train the model using that data.
% For pseudo data generation , we will use two strategies in this approach , which are as follows -
\subsubsection{Soft Labeling}
The intuition behind Soft Labeling is to reduce biased prediction by models for known intents. Thus, instead of using one hot label distribution, some part of known intent probability is reallocated to the open intent class which is known as Relocation probability ($\rho$). For learning these pseudo representations,  $KL$ divergence loss function \( L_{KL}\) is used.
    \begin{equation}
    \label{eqn:klloss}
    L_{KL}   = \sum_{i=1}^{N} p(s_i) ln(\frac{p(s_i)}{p'(s_i)})
\end{equation}
% \begin{equation}
%     \label{eqn:dkl}
%      D_{KL}(p(s_i)\parallel p'(s_i)) = p(s_i) ln(\frac{p(s_i)}{p'(s_i)})
% \end{equation}
where $N$ denotes size of training data, $p(s_i)$ denotes the softened probability distribution  of the user query \(s_i\) on
all intents, $p'(s_i)$ denotes output probability distribution when
softmax is applied on $\phi_{M+1}(e_i)$, and $e_i$  denotes query \(s_i\)  intent representation.
    
\subsubsection{Noisy Mixup}
This augmentation method combines noise injection and manifold mixup \cite{verma2019manifold} \cite{zhang2017mixup}, thereby including the benefits of both methods. Here, we assume that the pseudo data generated will belong to the class of open intent.
The method consist of following steps :-
\begin{itemize}
    \item Select a random hidden layer ($rl$)  of BERT.
    \item Select randomly two mini data batches $(S_{1},Y_{1})$ and $(S_{2},Y_{2})$ respectively. Filter out batches in such a way that instances at similar position in the batch belong to different intents.
    \item Filtered batches are then processed in the network till
    layer ($rl$).
    \item  The output of layer ($rl$) gives two intermediate mini-data batches on which Mixup and Noise injection is performed.
    % \begin{equation}
    % \begin{multlined}
    % O_{rl}(S_{1}) = BERT_{rl}(S_{1})\\
    % \noindent O_{rl}(S_{2}) = BERT_{rl}(S_{2})
    % \end{multlined}
    % \end{equation}
     \begin{equation}
    % \begin{multlined}
    O_{rl}(S_{1}) = BERT_{rl}(S_{1})
    % \\
    % \noindent O_{rl}(S_{2}) = BERT_{rl}(S_{2})
    % \end{multlined}
    \end{equation}
    \begin{equation}
        O_{rl}(S_{2}) = BERT_{rl}(S_{2})
    \end{equation}
    where \(  O_{rl}(S_{1})  \) and \(O_{rl}(S_{2}) \) represent the intermediate  mini data batches. 
    % and \(BERT_{rl}()\) represent random hidden layer \( rl \) of BERT.
    \item Perform Mixup on these intermediate mini data batches, producing the mixed mini data batch \( Mixup(S')\):
      \begin{equation}
        Mixup(S') = \lambda O_{rl}(S_{1}) + ( 1 - \lambda) O_{rl}(S_{2})
    \end{equation}
where the mixing level \(\lambda \sim Beta(\alpha, \alpha)\), with the hyper-parameters \(\alpha> 0\).
     \item Inject additive and multiplicative noise in the mixed mini data batch to produce noisy mixed mini data batch \(NM(S')\)
      \begin{equation}
      \scriptsize
        NM(S') = (1 + \delta_{mul} \times \xi_{mul} )Mixup(S') + \delta_{add}\times \xi_{add}
    \end{equation}
    where the \(\xi_{mul} \) and  \(\xi_{add}\) are independent random variables modeling the additive and multiplicative
noise respectively, and \(\delta _{mul} ,   \delta _{add}  \geq 0 \) are pre-specified noise levels.

    \item Continue forward pass from layer \( rl \) till the last layer of model.
    \ For learning these pseudo representations, softmax loss function \(L_{NM}\) is used.
\begin{equation}
     \label{eqn:noisymixuploss}
     L_{NM}  = -\frac{1}{N} \sum_{i=1}^{N}log{\frac{e^{(\phi_{M+1}(\Bar{{e_i}})^{M+1} )}}{\sum_{j=1}^{M+1}e^{(\phi_{M+1}(\Bar{{e_i}})^{j}) }}}
\end{equation}
where $\Bar{{e_i}}$ is the intent representation we get from last layer of model. $\phi_{M+1}(.)$
denote an ($M+1$)-class classifier. The output logits of the $j^{th}$ class are represented by $\phi_{M+1}(·)^j$. 

\end{itemize}

\begin{figure*}[htb]
\centering
\includegraphics [width=0.99\textwidth]{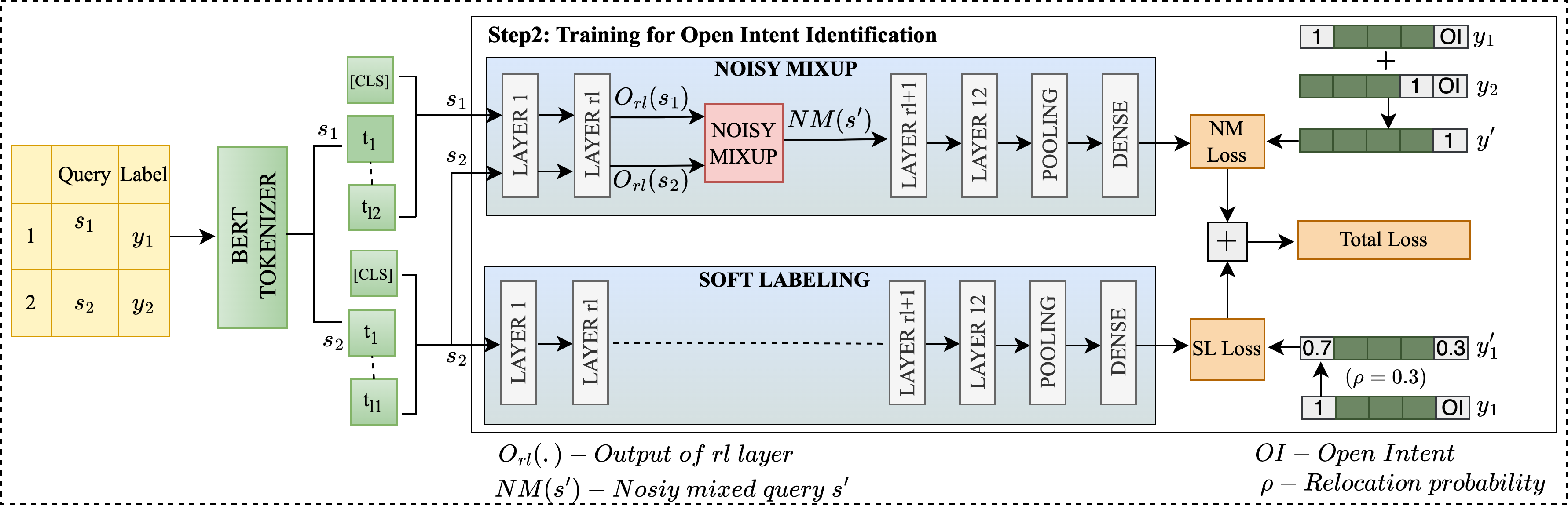}
\caption{\scriptsize It represents the functional view of SNOiC. For example, two intent instances  ($s_{1}$, $y_{1}$) and ($s_{2}$, $y_{2}$) (assuming mini-batch data size to 1 for simplicity) are passed to the BERT tokenizer for token generation. Two methods are used for model training: Soft Labeling and Noisy Mixup. Soft Labeling is achieved  by assigning  $\rho$ (default value is taken as 0.3) to the open intent class and $1-\rho$ (0.7) to the known intent class. The value $\rho$ is generally taken small to avoid overfitting the open intent class. The model learns through the total loss, combining the Soft labeling and Noisy Mixup losses.}
\label{fig:workflow}
\end{figure*}

\subsubsection{Loss Function} 
The loss function is the combination of Soft Labeling loss and Noisy Mixup loss shown in Equation \ref{eqn:TotalLoss}.  
\begin{equation}
\label{eqn:TotalLoss}
   L_{total} = \gamma L_{KL} + (1-\gamma)L_{NM}
\end{equation}
where \(\gamma\) can be either \( \lambda \) calculated above or can be a fixed tradeoff parameter. The value for \(\gamma\) varies according to the dataset and known intent class ratio $(r)$.

A working example of the proposed approach is shown in Figure 1.

\section{Experiments and Results}
This section first discusses the experimental setup, the datasets used and the performance metrics employed to evaluate the efficacy of our proposed SNOiC model. Next, it presents the results of our experiments, an ablation study and an analysis of the model's performance. 
% In this division, we describe our work's experimental settings, the datasets, and performance metrics used for evaluating the effectiveness of our proposed work.
% , and studying the number of performance metrics for our model.
\subsection{Experimental Setup}
The division of data for training, testing, and validation for the classifier is done per the previous studies \cite{shu2017doc,zhang2021deep,lin2019deep}.
% The datasets are divided into training , validation and testing sets . 
Some intent classes are kept as known for training, and the rest are kept open. The model is trained by keeping the known intent class ratio \((r)\) as 0.25, 0.5, and 0.75, and the rest classes are considered open. During testing both known and open intent classes are used. The BERT-base framework implemented in Pytorch is the backbone of our proposed model \cite{kenton2019bert}. The parameters for all the layers except the last are frozen to fasten the training procedure. In soft labeling, the default value of relocation probability ($\rho$) is $0.3$ and is kept fixed for all permutations and combinations. The hyperparameter \(\alpha\) and \(\gamma\) vary according to the dataset and known intent class ratio $(r)$. The default values for additive and multiplicative noise levels ($\delta_{add}$ and $\delta_{mul}$) are $0.4$ and $0.2$, respectively.  
The optimizer used is AdamW~\cite{loshchilov2017decoupled}, and the learning rate is set to $2e-5$. A batch size of $128$ is used for the training process. During the training process, the model undergoes 100 epochs, and the best-performing model is chosen by evaluating its performance on the validation set with early stopping. The source code is available at GitHub repository \footnote{\#link is omitted for anonymity} (link is omitted for anonymity) and uploaded during the paper submission.

~\vspace{-3.0ex}

\subsection{Datasets}
Our proposed SNOiC model's efficacy was assessed by conducting experiments on four benchmark datasets.
\begin{table}[htb]
 \centering
\caption{Dataset details}
\resizebox{\linewidth}{!}{%
\begin{tabular} {c|c|c|c|c|c|c}
 \hline
 \textbf{Dataset} & \textbf{Classes} & \textbf{Train} & \textbf{Validation} & \textbf{Test} & \textbf{Vocab Size} & \textbf{Mean Length} \\
 \hline
 BANKING & 77  & 9003 & 1000 & 3080 & 5028 & 11.91  \\
\hline
 CLINIC & 150  & 15000 & 3000 & 5700 & 8376 & 8.31  \\
\hline
 ATIS & 7  & 13084 & 700 & 700 & 11971 & 9.05  \\
\hline
 SNIPS & 18  & 4978 & 500 & 893 & 938 & 11.37  \\
\hline
\end{tabular}}

\label{tab:Dataset}
\end{table}
The BANKING dataset comprises 13,083 customer service queries in the banking domain, which are categorized into 77 unique intents \cite{casanueva2020efficient}. The CLINIC dataset spans ten domains, encompassing a total of 22,500 in-scope queries categorized into 150 distinct intents, along with 1,200 out-of-scope queries \cite{larson2019evaluation}. The ATIS dataset is centered around the airline travel domain and consists of 4,978 queries categorized into 18 distinct intents \cite{hemphill1990atis}. The SNIPS dataset encompasses seven distinct intents spanning across multiple domains \cite{coucke2018snips}. The complete details of the dataset used is shown in Table \ref{tab:Dataset}.

   %  \noindent\textbf{BANKING} It is a dataset comprising 13,083 customer service queries in the banking domain, which are categorized into 77 unique intents \cite{casanueva2020efficient}.\\
   %  \noindent\textbf{CLINIC} The CLINIC dataset spans ten domains, encompassing a total of 22,500 in-scope queries that are categorized into 150 distinct intents, along with 1,200 out-of-scope queries \cite{larson2019evaluation}.\\
   %  \noindent\textbf{ATIS} This dataset is centered around the airline travel domain, consisting of 4,978 queries that are categorized into 18 distinct intents \cite{hemphill1990atis}.\\
   % \noindent\textbf{SNIPS} It encompasses seven distinct intents, spanning across multiple domains \cite{coucke2018snips}.

% \vspace{-3.0ex}

% \vspace{-4.0ex}
\subsection{Performance Metrics}
The metrics used for evaluating our SNOiC model's effectiveness such as Accuracy, Recall, Precision, which are in accordance with previous studies~\cite{shu2017doc,zhang2021deep,lin2019deep}.

    % \item \textbf{Accuracy} : \textcolor{red}{Accuracy score is calculated using direct accuracy\_score function from Scikit Learn Library.}
    \noindent\textbf{\textit{F1-score} on all classes:} To calculate the \textit{F1-score} for all classes, we first find the \textit{F1-score} for each class by taking the harmonic mean of precision and recall. After that, we compute the average of all the \textit{F1-scores} obtained from each class to get the final \textit{F1-score} value.
    % \begin{itemize}
    %     \item Fist F1-score over each individual class is computed. F1-score over each individual class is the harmonic mean of precision and recall, two other commonly used evaluation metrics. Precision measures the proportion of true positives (correctly predicted positive samples) out of all positive predictions. Recall measures the proportion of true positives out of all actual positive samples.
    %     \item Then the average of the F1-scores for all classes is computed for the final value.
    % \end{itemize} 
    % F1-score is calculated as follow :
    
    \begin{equation}
       \text{\textit{F1-score}} = \frac{1}{M+1}\sum_{a=1}^{M+1}2 \times \frac{P_{C_a} \times R_{C_a}}{P_{C_a} + R_{C_a}}
    \end{equation}
    \begin{equation}
    \label{eqn:precision}
        P_{C_a} =\frac{TP_{C_a}}{TP_{C_a} + FP_{C_a}}
    \end{equation}
    \begin{equation}
    \label{eqn:recall}
        R_{C_a} = \frac{TP_{C_a}}{TP_{C_a} + FN_{C_a}}
    \end{equation}
where $P$ and $R$ represent Precision, and the Recall score over $M+1$ classes, $P_{C_a}$ and $R_{C_a}$, stands for precision and recall score on the $C_a$ class, $TP_{C_a}$, $FP_{C_a}$, and $FN_{C_a}$ stand for true positives, false positives, and false negatives of the $C_{a}$ class, respectively.

\noindent\textbf{Macro F1-score over Known Intent classes ($F1_{KI}$) and Open Intent class $(F1_{OI})$} : $F1_{KI}$ measures the model's performance across all known intent classes by computing the F1-score for each class and then taking the average of these \textit{F1-scores}. $F1_{OI}$ computes \textit{F1-score} only over open intent class. They are calculated as follows:
    \begin{equation}
        F1_{KI}\text{\textit{-score}}  = \frac{1}{M}\sum_{a=1}^{M}2 \times \frac{P_{C_a} \times R_{C_a}}{P_{C_a} + R_{C_a}}
    \end{equation}
    \begin{equation}
        F1_{OI}\text{\textit{-score}} = 2\times \frac{P_{C_{M+1}}\times R_{C_{M+1}}}{P_{C_{M+1}}+R_{C_{M+1}}}
    \end{equation}
        
where $F1_{KI}$ stands for \textit{F1-score} over known intent classes and  $F1_{OI}$ stands for \textit{F1-score} over open intent class, $P_{C_a}$  
 (Equation \ref{eqn:precision} ) and $R_{C_a}$ (Equation \ref{eqn:recall}), stands for precision and recall score on the $C_a$ class, $P_{C_{M+1}}$ stand for precision over open intent class and  $R_{C_{M+1}}$ stand for recall over open intent class.    

\subsection{Results}
 We conducted experiments on four benchmark datasets with 25\%, 50\%, and 75\% known intent class ratios $(r)$, respectively, and reported accuracy, F1-score, Known \textit{F1-score} ($F1_{KI}$-score), and Open \textit{F1-score} ($F1_{OI}$-score) to validate the results. Table \ref{table:results} summarizes the results of our proposed SNOiC model. As it can be observed, the CLINIC dataset achieves the highest performance on all four performance metrics
because it is a balanced dataset with the highest number of intent instances and classes. The BANKING is a fine-grained banking domain dataset, due to which model performs pretty well on this dataset. 

In the case of ATIS, which is an imbalanced dataset, the results are comparatively poor when the $r$ is 25\%. However, the results improve with an increase in the value of $r$. Similarly, in the SNIP dataset, a decline in performance can be observed.
% In case of ATIS, it is an imbalanced dataset due to which it performs poorly when known class ratio is 25\% and increases with increase in known class ratio. Similarly, a decline in performance can be observed in the SNIP dataset. 
The reason could be fewer classes, which are even less than 10. When $r$ is 25\%, the number of known intent classes drops relatively low, leading to decreased performance. An increase in the number of classes is directly proportional to a rise in the model's performance.
 % \vspace{-4.0ex}
\begin{table}[h!]
\centering
% \scriptsize
\caption{Experimental results of SNOiC model with different known intent class ratio values $r$}
\label{table:results}
\resizebox{\linewidth}{!}{%
\begin{tabular} {p{3.0cm}|c|c|c|c|c}
 \hline
 \textbf{Known intent class ratio $(r)$} & \textbf{Dataset} & \textbf{Accuracy} & \textbf{F1-score} & \textbf{$F1_{KI}$-score}  & \textbf{$F1_{OI}$-score}   \\
 \hline
\multirow{4}{4em}{\centering 0.25} & BANKING   & 82.45&	72.89&	72.11&	87.75  \\

 & CLINIC  & 91.61	&81.54&	81.20&	94.71  \\

 & ATIS  & 81.78	& 65.97&	65.02&	69.20 \\

  & SNIPS & 72.10&	71.11&	68.23&	76.86  \\
\hline
\multirow{4}{4em}{\centering 0.5} & BANKING   & 80.33&	82.10&	82.15&	80.23  \\

 & CLINIC  & 88.90&	86.85&	86.80&	91.10 \\

 & ATIS  & 91.14&	80.54&	82.04&	70.30 \\

  & SNIPS & 78.72&	82.98&	86.47&	69.02  \\
\hline
\multirow{4}{4em}{\centering 0.75} & BANKING   & 81.98	&86.58&	86.88&	68.72 \\

 & CLINIC  & 88.21&	89.77&	89.80&	86.47  \\

 & ATIS  & 95.69&	84.54&	85.43	&74.17 \\

  & SNIPS & 85.91&	88.20&	91.63&	71.00 \\
\hline
\end{tabular}}
\end{table}

% \vspace{-1.0ex}
\begin{table*}[htb]
\centering
% \scriptsize
\caption{Experimental analysis with different variants of SNOiC models}
\label{tab:AblationSTUDY}
\tiny
\begin{tabular}{c|c|cccc|cccc}
\hline
\textbf{}             &           & \multicolumn{4}{c|}{\textbf{Dataset - SNIPS}}  & \multicolumn{4}{c}{\textbf{Dataset - BANKING}} \\ \hline
$r$                 & Model   & Accuracy    & F1-score  & $F1_{KI}$-score  & $F1_{OI}$-score  & Accuracy   & F1-score   & $F1_{KI}$-score  & $F1_{OI}$-score  \\ \hline
\multirow{4}{*}{0.25} & SNOiC-SL  & 45.44       & 55.28     & 64.78     & 36.28    & 47.00      & 53.17      & 53.58     & 45.42    \\
                      & SNOiC-MN  & 61.63       & 68.32     &   \textbf{72.09}     & 60.79    & 81.53      & 72.73      & 71.98     & 86.98    \\
                      & SNOiC-AN  & 62.37       & 68.80     & 71.96     & 62.46    & 82.10      & 72.89      & 72.11     & 87.50    \\
                      & SNOiC       &   \textbf{72.10}       &   \textbf{71.11}     & 68.23     &   \textbf{76.86 }   &   \textbf{82.45}      &   \textbf{72.89}      &   \textbf{72.11 }    &   \textbf{87.75}  \\ \hline
\multirow{4}{*}{0.5}  & SNOiC-SL  & 66.13       & 71.94     & 81.00        & 35.66    & 54.36      & 68.81      & 69.93     & 26.00       \\
                      & SNOiC-MN  & 77.68       & 82.06      & 85.99     & 66.34    & 80.23      & 82.09      & 82.15     & 80.02    \\
                      & SNOiC-AN  &   \textbf{78.99}       &   \textbf{83.09 }    & \textbf{86.47 }     &   69.46    & 80.23      & 82.09     & 82.15     & 80.02    \\
                      & SNOiC       & 78.72       & 82.98     &   \textbf{86.47}     & 69.02    &   \textbf{80.33}   &   \textbf{82.10 }     &  \textbf{ 82.15}    &   \textbf{80.23}    \\ \hline
\multirow{4}{*}{0.75} & SNOiC-SL  & 73.50       & 81.75     &  81.00         &    35.66      & 72.98      & 82.05      & 83.10     & 20.87    \\
                      & SNOiC-MN  & 82.92       & 85.66     & 90.61      & 60.92    & 81.98      & 86.56      & 86.88     & 68.72    \\
                      & SNOiC-AN  & 83.15       & 86.41     & 90.75     & 63.72    & 81.98      & 86.56      & 86.88     & \textbf{68.72 }   \\
                      & SNOiC       &   \textbf{85.91 }      &   \textbf{88.20}     &   \textbf{91.63}     &   \textbf{71.00}    &   \textbf{81.98}     &   \textbf{86.58}    &   \textbf{86.88}   &   \textbf{68.72}   \\ \hline
\textbf{}             & \textbf{} & \multicolumn{4}{c|}{\textbf{Dataset - CLINIC}} & \multicolumn{4}{c}{\textbf{Dataset - ATIS}}    \\ \hline
$r$                 & Model   & Accuracy    & F1-score  & $F1_{KI}$-score  & $F1_{OI}$-score  & Accuracy   & F1-score   & $F1_{KI}$-score  & $F1_{OI}$-score  \\ \hline
\multirow{4}{*}{0.25} & SNOiC-SL  & 61.76       & 56.73     & 56.42     & 68.29    & 75.70      & 69.73      & 61.68     & 42.56    \\
                      & SNOiC-MN  & 90.70       & 81.30     & 80.97     & 94.07    & 79.93      & 58.85      & 58.37     & 58.31    \\
                      & SNOiC-AN  & 90.57       & 81.12     & 80.78     & 93.98    & 80.00      & 50.66      & 47.23     & 61.30    \\
                      & SNOiC       &  \textbf{ 91.61  }     &   \textbf{81.54}     &   \textbf{81.20 }    &   \textbf{94.71}    &   \textbf{81.78}      &   \textbf{65.97}      &   \textbf{65.02}     &  \textbf{ 69.20 }   \\ \hline
\multirow{4}{*}{0.5}  & SNOiC-SL  & 64.97       & 71.78     & 71.93     & 60.82    & 65.96      & 71.31      &   76.01        & 23.34         \\
                      & SNOiC-MN  & 88.73       & 86.73     & 86.68     & 90.72    & 90.27      & 77.05      & 78.53     & 66.76    \\
                      & SNOiC-AN  & 79.68       & 83.56     & 86.75     & 70.90    & 90.27      & 77.05      & 78.53     & 66.76    \\
                      & SNOiC       &  \textbf{ 88.90 }      &   \textbf{86.85}     &  \textbf{ 86.80}     &   \textbf{91.10}    &   \textbf{91.14}      &   \textbf{80.54}      &  \textbf{ 82.04 }    &  \textbf{ 70.30 }    \\ \hline
\multirow{4}{*}{0.75} & SNOiC-SL  & 73.50       & 81.75     & 81.96     & 57.95    & 86.68      & 78.37      & 82.44     & 30.74    \\
                      & SNOiC-MN  & 88.20       & 89.80     & 89.83     & 86.42    & 95.52      & 83.93      & 84.84     & 73.35    \\
                      & SNOiC-AN  &   \textbf{88.30 }      &   \textbf{89.82}     &   \textbf{89.85}  &   \textbf{86.60}    & 95.52      & 83.93      & 84.84     & 73.35    \\
                      & SNOiC       & 88.21       & 89.77     & 89.80      & 86.47    &   \textbf{95.69}     &   \textbf{84.54}     &  \textbf{ 85.43}     &   \textbf{74.17}   \\ \hline
\end{tabular}

\end{table*}

\subsection{Ablation Study}

In this section, an ablation study is performed to examine the effectiveness of the proposed SNOiC model by conducting various experiments. For this, we created three variants of the SNOiC Model i.e., SNOiC without Soft Labeling (SNOiC-SL), SNOiC without injecting Additive Noise (SNOiC-AN), and SNOiC without injecting Multiplicative Noise (SNOiC-MN). The experiments were conducted on all four datasets. We calculated Accuracy and \textit{F1-score} for all the four variants while maintaining all other parameters same. Table \ref{tab:AblationSTUDY}  shows the experimental analysis with different SNOiC models. The findings are as follows:

% \usepackage{multirow}
% \usepackage{multirow}

\begin{itemize}
    \item The value of performance metrics decreases
for all three variants of the SNOiC Model, i.e., SNOiC-SL, SNOiC-AN, SNOiC-MN, indicating that all components collectively 
contribute toward the performance of the model.
\item It can be observed that for the variant SNOiC-SL, the accuracy and \textit{F1-score} drop by at least 10\% in all the cases, which shows that soft labeling is quite efficient. Furthermore, a massive decrease in $F1_{OI}$ can be observed as the model tends to get biased towards known intent classes. For the CLINIC dataset, when the $r$ is 75\%, the accuracy decreases from 88.21\% to 73.50\%, and $F1_{OI}$ decreases from 86.47\% to 57.95\%, which shows that in all cases, whether labeled data is in large or small amounts, removing soft labeling has a massive impact on the model.

\item The value of the performance for variant SNOiC-MN declines more than the variant SNOiC-AN. Noise is injected to cover a large area of data surroundings, and injecting
multiplicative noise has a more significant role which can be seen from the results. A significant decrease in performance metrics values can be observed for SNOiC-AN and SNOiC-MN variants for the SNIPS dataset's when $r$ is 25\%. The accuracy decreases from 72.10\% to 62.37\% and 61.63\%, respectively, and $F1_{OI}$ also decreases from 76.86\% to 62.46\% and 60.79\%, respectively, for SNOiC-AN and SNOiC-MN variants.
% \item A marginal decrease in performance for variants  SNOiC-AN and SNOiC-MN  can be observed. 
% \textcolor{red}{This happens due to the class distribution properties of the BANKING dataset.} Also, the same trend can be observed in the CLINIC dataset.

\end{itemize}

\subsection{Model Analysis}
In this section, we study the effect of different parameters on our proposed SNOiC model by varying the values of the following parameters: Relocation Probability $(\rho)$, Labeled data ratio, and Noise Levels, and studied their effect on our SNOiC model.

\noindent\textbf{Effect of Relocation Probability}: We analyzed the effect of varying the relocation probability  $\rho$ from 0.1 to 0.5. We conducted experiments on ATIS and CLINIC datasets with the $r$ as 25\%  and analyzed performance using Accuracy and \textit{F1-score}.
    As can be observe from Figure \ref{1}, Accuracy and \textit{F1-score} for both datasets first increase and then decrease. The accuracy and \textit{F1-score} are the highest in the CLINIC dataset when  $\rho$ is 0.3. The ATIS dataset's \textit{F1-score} is highest when  $\rho$ is 0.3 and then decreases sharply.
    As  $\rho$ increases, the model gets more biased towards the open intent class and does not accurately classify known intent classes, resulting in decreasing performance.

\begin{figure}[htb]%
    \centering
    \subfloat[\centering ATIS (25\% known intents)]{{\includegraphics[width=5.5cm]{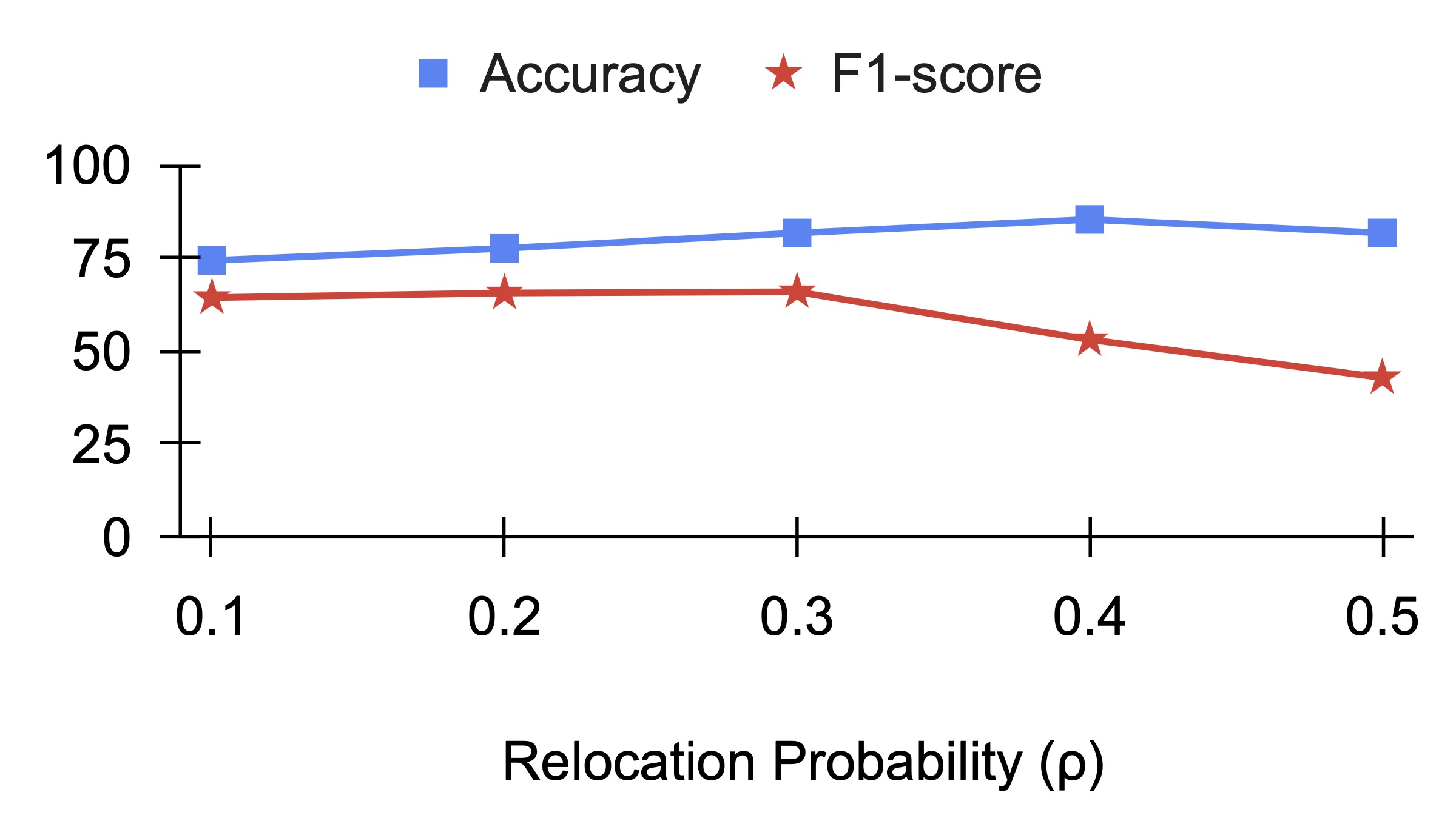} }}%
    \qquad
    \subfloat[\centering CLINIC (25\% known intents)]{{\includegraphics[width=5.5cm]{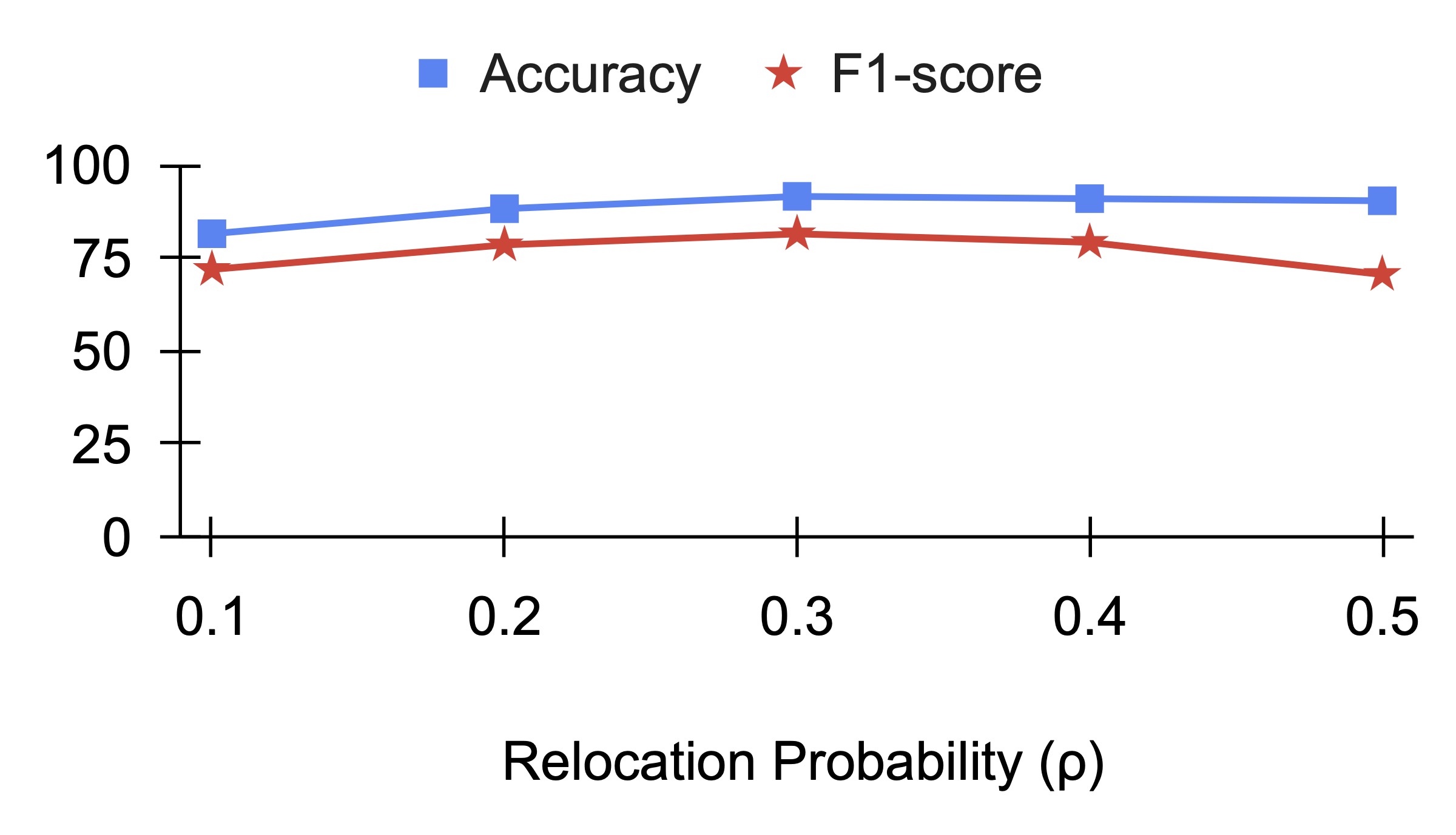} }}%
    \caption{Effect of Relocation Probability by varying  $\rho$ on ATIS and CLINIC dataset with $r$ set as 25\%.}%
    \label{fig:Labelsl}%
\end{figure}

\begin{figure*}[htb]
\centering
 \includegraphics[width=0.99\textwidth]{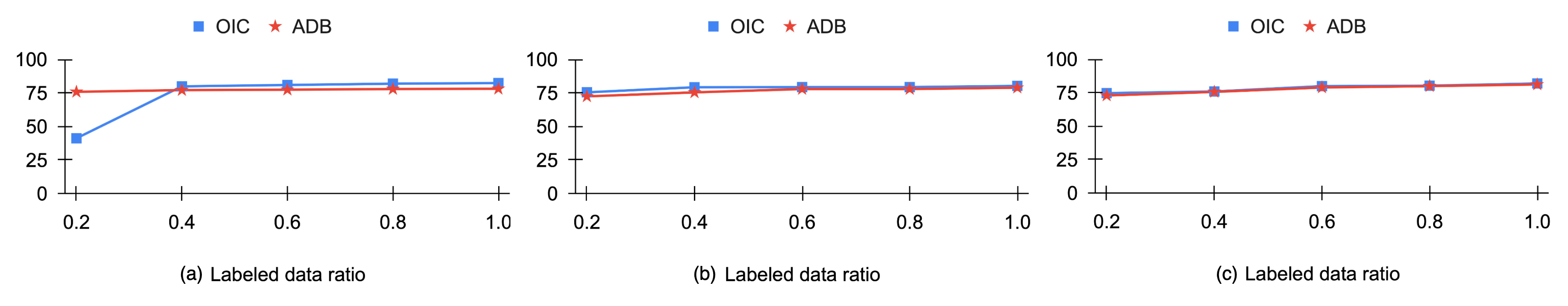}
\caption{Effect of labeled data ratio on BANKING dataset with different values of $r$, (a) $r = 25\%$, (b) $r = 50\%$, (c) $r = 75\%$.}
\label{2}
\end{figure*}

\begin{table*}[]
\caption{Effect of increased noise levels on four datasets with different values of $r$}
\label{table:noisestudy}
\tiny
\begin{tabular}{c|c|cc|cc|cc|cc}
\hline
\multirow{2}{*}{$r$ \& $\alpha$}                                                 & \multirow{2}{*}{Noise Levels} & \multicolumn{2}{c}{\textbf{BANKING}}    & \multicolumn{2}{c}{\textbf{CLINIC}}     & \multicolumn{2}{c}{\textbf{ATIS}}       & \multicolumn{2}{c}{\textbf{SNIPS}}      \\ \cline{3-10} 
                                                                        &                               & \multicolumn{1}{c|}{Accuracy} & F1-score & \multicolumn{1}{c|}{Accuracy} & F1-score & \multicolumn{1}{c|}{Accuracy} & F1-score & \multicolumn{1}{c|}{Accuracy} & F1-score \\ \hline
\multirow{2}{*}{\begin{tabular}[c]{@{}c@{}} $r$= 0.25\\ $\alpha = 2$ \end{tabular}} & $\delta_{add}$ = 0.4, $\delta_{mul}$ = 0.2          & \textbf{75.53}    & \textbf{70.88 }   & \textbf{90.75}   & \textbf{81.28}    & \textbf{80.20}    & \textbf{71.63}    & \textbf{69.43}    & 63.20    \\ \cline{2-10} 
                                                                        & $\delta_{add}$ = 0.8, $\delta_{mul}$ = 0.4          & 74.53    & 70.88    & 89.93    & 80.62    & 77.68    & 70.48    & 63.20    & \textbf{69.45 }   \\ \hline
\multirow{2}{*}{\begin{tabular}[c]{@{}c@{}} $r$= 0.5\\ $\alpha = 2$ \end{tabular}}  & $\delta_{add}$ = 0.4, $\delta_{mul}$ = 0.2          & \textbf{80.30}    & \textbf{82.09}    & \textbf{88.90}    & \textbf{87.00 }   & \textbf{91.94}    & \textbf{80.54}    & \textbf{78.54}    & \textbf{82.84}    \\ \cline{2-10} 
                                                                        & $\delta_{add}$ = 0.8, $\delta_{mul}$ = 0.4          & 80.30    & 82.09    & 88.67    & 86.78    & 90.27    & 77.05    & 78.54    & 82.84    \\ \hline
\multirow{2}{*}{\begin{tabular}[c]{@{}c@{}}$r$= 0.75\\ $\alpha = 2$ \end{tabular}} & $\delta_{add}$ = 0.4, $\delta_{mul}$ = 0.2          & \textbf{81.97}    & \textbf{86.57}    & \textbf{88.21 }   & \textbf{89.77 }   & \textbf{95.69 }   & \textbf{84.54}    & \textbf{82.90 }   & \textbf{85.70 }   \\ \cline{2-10} 
                                                                        & $\delta_{add}$ = 0.8, $\delta_{mul}$ = 0.4          & 81.97    & 86.57    & 88.21    & 89.77    & 95.52    & 83.93    & 82.90    & 85.70    \\ \hline
\end{tabular}
\end{table*}

% \begin{figure}
% \centering
% \begin{minipage}{.5\textwidth}
%   \centering
%   \includegraphics[width=.99\linewidth]{ATIS-25.png}
%   \captionof{figure}{A figure}
%   \label{fig:test1}
% \end{minipage}%
% \begin{minipage}{.5\textwidth}
%   \centering
%   \includegraphics[width=.99\linewidth]{CLINIC-25.png}
%   \captionof{figure}{Effect of Soft Labeling by varying probability allocated to open intent class}
%   \label{fig:test2}
% \end{minipage}
% \end{figure}

\noindent\textbf{Effect of Labeled data ratio}: We analyzed the effect of the varying ratio of labeled data on our proposed SNOiC model. We conducted experiments on the BANKING dataset with varying $r$ as 25\%, 50\%, and 75\%  and compared our result with another model ADB~\cite{zhang2021deep}. As shown in Figure \ref{2}, our model only underperform  ADB when the $r$ is 0.25 and labeled data ratio is 0.2, and for the rest permutations and combinations, it performs pretty well. The reason being the pseudo data generated is relatively low, and the classifier cannot classify data for the open intent class. With an increase in data, the accuracy of the classifier also increases, showing that it performs well even with increased data.

% \begin{figure}[htp]
% \centering
% \includegraphics[width=.3\textwidth]{Banking-25.png}\hfill
% \includegraphics[width=.3\textwidth]{Banking-50.png}\hfill
% \includegraphics[width=.3\textwidth]{Banking-75.png}
% \caption{default}
% \label{fig:figure3}

% \end{figure}
\noindent\textbf{Effect of Noise Levels}: We analyzed the effect on the classifier when noise levels are increased. We conduct experiments on all four datasets and fix $\alpha$ as 2. Also, two sets of noise levels are used, one is $\delta_{add}$ = 0.4, $\delta_{mul}$ = 0.2, and the other is $\delta_{add}$ = 0.8, $\delta_{mul}$ = 0.4.
It can be observed from the Table \ref{table:noisestudy} that with noise levels as $\delta_{add}$ = 0.4, $\delta_{mul}$ = 0.2, we achieve better results. Increased noise levels increase the natural region surrounding the classes and could lead to overlapping, resulting in decreased performance.

\begin{table*}[htb]
\caption{Performance comparison with different state-of-the-arts on two benchmark datasets}
\tiny
\label{tab:Comparison}

\begin{tabular}{l|l|cccc|cccc} 
 \hline
               &                & \multicolumn{4}{c|}{\textbf{Dataset - BANKING}}                                                                                                               & \multicolumn{4}{c}{\textbf{Dataset - CLINIC}}                                                                                                                \\  \hline
\textbf{$r$} & \textbf{Model} & \textbf{Accuracy}& \textbf{F1-Score} & \textbf{$F1_{KI}$-score} & \textbf{$F1_{OI}$-score} & \textbf{Accuracy} & \textbf{F1-Score} & \textbf{$F1_{KI}$-score} & \textbf{$F1_{OI}$-score} \\  \hline
          & DOC            & 56.99                                 & 58.03                                 & 57.85                                 & 61.42                                & 74.97                                 & 66.37                                 & 65.96                                 & 81.98                                \\
               & Openmax        & 49.94                                 & 54.14                                 & 54.28                                 & 51.32                                & 68.5                                  & 61.99                                 & 61.62                                 & 75.76                                \\
       0.25         & LMLC           & 64.21                                 & 61.36                                 & 60.88                                 & 70.44                                & 81.43                                 & 71.16                                 & 70.73                                 & 87.33                                \\
              
               & KNNCL            & 73.01                                 & 66.23                                 & 65.54                                 & 79.34                                & 89.87                                 & 79.23                                 & 78.85                                 & 93.56                                \\
                & ARPL            & 76.80                                 & 64.01                                 & 62.99                                 & 83.39                                & 84.51                                 & 73.44                                 & 73.01                                 & 89.63                                \\            
                & ADB            & 78.18                                 & 70.54                                 & 69.82                                 & 84.08                                & 87.79                                 & 77.63                                 & 77.27                                 & 92.01                                \\
                & DA-ADB           & 81.19                                 & \textbf{73.73}                                 &   \textbf{73.05}                               &       86.57                          & 89.48                                 & 79.92                                 &                  79.57                &                      93.20           \\
               & SNOiC            &  \textbf{ 82.45}                             &   72.89                                &   72.11                                 &   \textbf{87.75}                               &   \textbf{91.61}                                 &   \textbf{81.54 }                                &   \textbf{81.20 }                                 &   \textbf{94.71}                                \\  \hline
            & DOC            & 64.81                                 & 73.12                                 & 73.59                                 & 55.14                                & 77.16                                 & 78.26                                 & 78.25                                 & 79.00                                   \\
               & Openmax        & 65.31                                 & 74.24                                 & 74.76                                 & 54.33                                & 80.11                                 & 80.56                                 & 80.54                                 & 81.89                                \\
   0.5            & LMLC           & 72.73                                 & 77.53                                 & 77.74                                 & 69.53                                & 83.35                                 & 82.16                                 & 82.11                                 & 85.85                                \\

                & KNNCL            & 70.41                                    & 74.96                                 & 75.16                                 & 67.21                                & 85.32                                 & 83.31                                 & 83.25                                    & 87.85                                \\
                & ARPL            & 74.11                                 & 77.77                                 & 77.93                                 & 71.79                                & 80.36                        & 80.88                                 & 80.87                                 & 81.81                                \\ 

                 & ADB            & 79.00                                    & 80.88                                 & 80.93                                 & 78.71                                & 86.37                                 & 85.04                                 & 85.00                                    & 88.46                                \\
               
               & DA-ADB           &\textbf{ 81.51}                                 & 82.53                                 &   82.54                                &    \textbf{81.93}                            & 87.93                                 & 85.64                                 &                 85.58                &                              90.10  \\
               
               & SNOiC            &   80.33                                 &   \textbf{82.80}                                  &   \textbf{82.85}                                &   80.23                                &   \textbf{88.90}                                  &  \textbf{ 86.85}                                 &   \textbf{86.80 }                                &   \textbf{91.10 }                                \\  \hline
           & DOC            & 76.70                                  & 83.34                                 & 83.91                                 & 50.60                                 & 78.73                                 & 83.59                                 & 83.69                                 & 72.87                                \\
               & Openmax        & 77.45                                 & 84.07                                 & 84.64                                 & 50.85                                & 76.8                                  & 73.16                                 & 73.13                                 & 76.35                                \\
   0.75            & LMLC           & 78.52                                 & 84.31                                 & 84.64                                 & 58.54                                & 83.71                                 & 86.23                                 & 86.27                                 & 81.15                                \\

                & KNNCL            & 74.78                                 & 81.25                                 & 81.76                                & 51.42                                & 84.12                                 & 86.10                                    & 86.14                                 & 82.05                                 \\
                 & ARPL            & 79.60                                 & 85.16                                 & 85.58                                 & 61.26                                & 81.29                        & 86.00                                 & 86.10                                 & 74.67                                \\ 
                & ADB            & 81.26                                 & 86.05                                 & 86.37                                 & 67.08                                & 87.05                                 & 89.00                                    & 89.02                                 & 84.90                                 \\

               & DA-ADB           & 81.12                                 & 85.65                                 &   85.93                               &      \textbf{69.37}                           & 87.39                                 & 88.41                                 &                 88.43                  &               86.00                 \\
               & SNOiC            &   \textbf{81.98}                                 &   \textbf{86.58 }                                &   \textbf{86.88}                                 &  68.72                                &   \textbf{88.21 }                                &  \textbf{89.77 }                                &   \textbf{89.80}                                  &   \textbf{86.47}                                \\  \hline
\end{tabular}

{ $*$ The results for ADB~\cite{zhang2021deep} and DA-ADB~\cite{zhang2023learning}, KNNCL~\cite{zhou2022knn}, ARPL~\cite{chen2021adversarial} models are calculated using open source code~\footnote{https://github.com/thuiar/TEXTOIR/tree/main/open_intent_detection}. For rest of models DOC, Openmax and LMCL, results were taken from paper~\cite{zhang2021deep}.}
\vspace{-3.0ex}
\end{table*}

\subsection{Comparison with State-of-the-art Methods }
This section compares our proposed SNOiC model with state-of-the-art models like DOC~\cite{shu2017doc}, OpenMax~\cite{bendale2016towards}, LMCL~\cite{lin2019deep}, KNNCL~\cite{zhou2022knn}, ARPL~\cite{chen2021adversarial}, ADB~\cite{zhang2021deep}, DA-ADB~\cite{zhang2023learning}. Table \ref{tab:Comparison} compares performance using  metrics Accuracy, F1-score, $F1_{KI}$  and $F1_{OI}$.

% Table \ref{table:comparison} and Table \ref{table:comparison1} shows the performance of our model with respect to other state-of-the-art methods.  \\
Table \ref{tab:Comparison} illustrates that the SNOiC model outperforms all other state-of-the-art methods in most cases. Compared to DA-ADB on the CLINIC dataset, our SNOiC model outperforms all the measures with an $r$ value of 25\%, 50\%, and 75\%, respectively. Similarly, compared to DA-ADB on the BANKING dataset, our SNOiC model outperforms most measures with an $r$ value of 25\%, 50\%, and 75\%, respectively. Specifically, our model surpasses others in terms of accuracy and F1-score metrics, demonstrating its effectiveness in detecting open intent classes and accurately classifying known ones. Additionally, even with limited labeled data (25\% $r$ value), our method performs exceptionally well. The results clearly indicate a notable improvement in the performance of the SNOiC model compared to other state-of-the-art models.

% Specifically, in terms of the accuracy metric, when compared to ADB, our SNOiC model surpasses others by 4.27\%, 1.33\%, and 0.72\% on the BANKING dataset and by 3.82\%, 2.5\%, and 1.16\% on the CLINIC dataset with an $r$ value of 25\%, 50\%, and 75\%, respectively. Likewise, in terms of the F1-score metric and compared to ADB, our method outperforms it by 2.35\%, 1.22\%, and 0.53\% on the BANKING dataset and by 3.91\%, 1.81\%, and 0.77\% on the CLINIC dataset with an $r$ value of 25\%, 50\%, and 75\%, respectively. Additionally, improvement is observed when the $r$ value is 25\%, indicating that our method performs exceptionally well when labeled data is limited. The results indicate a notable improvement in the performance of SNOiC compared to other state-of-the-art models, demonstrating its effectiveness in detecting open intent classes and accurately classifying known ones. When compared with ADB on metrics $F1_{KI}$ and $F1_{OI}$, our model's performance increases by 2.29\%, 1.22\%, and 0.51\% for known intent classes and 3.67\%, 1.53\%, and 1.64\% for open intent classes on the BANKING dataset with 25\%, 50\%, and 75\% settings, respectively.

% \usepackage{multirow}

\subsection{Limitations}
% In this paper , we have proposed a model for Open Intent Classification. The problem can generally be seen as a M + 1 classification problem with M known classes and one open intent class.
% In the past, open intent identification has been performed using methods based on thresholds. However, these methods are susceptible to overfitting and can lead to overly confident predictions, which can make it challenging to establish an appropriate threshold.
% Also, the unavailability of open intent data poses a major challenge while training model for M+1 class. To mitigate the above limitations, we have trained our model using two methods soft labeling and Noisy Mixup. By reallocating probability from known intent classes to open intent classes, Soft Labeling mitigates the issue of overconfident predictions in the model. Meanwhile, Noisy Mixup, a data augmentation technique tackles the lack of open intent data by creating pseudo samples for training M+1 class .
% By performing experiments, we found out that our model achieves better results on all four performance metrics when compared with the other state-of-art-models. 
From the experiments, it was observed that the performance of SNOiC for the CLINIC dataset was higher than the ATIS dataset for performance metrics $F1_{KI}$ and $F1_{OI}$. The SNOiC's performance suffers when training with an imbalanced dataset, which poses a limitation. One way to mitigate this issue could be to apply a method for balancing the data. It was also observed that the model's performance decreases when the number of classes is significantly low, as is the case with SNIPS, where $r$ is 25\% (known intent classes are between 2 to 3). This dip in performance is noticeable when compared to other datasets and can be seen as a limitation of the model. These limitations can be addressed in future work.
% It was also observed that model's performance  dips when the number of classes are very less i.e in case of SNIPS when known ratio is 25\% ( known classes lie between [2-3]) dip in numbers when compared with other datasets can be observed which can be viewed as model, and can be viewed as another limitation of the model. The above limitations can be handled in future work.
\section{Conclusion}
In this paper, we proposed SNOiC for open intent classification. The model utilizes two methods, Soft labeling, and Noisy Mixup. With soft labeling, each sample is assigned a probability of being predicted as an open intent. At the same time, Noisy Mixup generates pseudo-open intent samples through a combination of Mixup and Noise Injection. Our model performs ($M+1$) class classification by employing these two approaches. We evaluated our proposed model on four benchmark datasets through extensive experiments, and the results demonstrate that the SNOiC model achieved the maximum and minimum performance of 94.71\% and 68.72\% while identifying the open intents. Also, the SNOiC model improved the performance of identifying the open intents by 12.76\% (max) and 0.93\% (min) compared with state-of-the-art models. It has been observed that the performance of the SNOiC depends on the characteristics of the dataset. The proposed model will be helpful for the research community to develop more robust dialog-based systems.  In the future, we will explore class incremental learning (CIL) strategies to update the SNOiC with identified open intents.
\bibliographystyle{acl_natbib}
\bibliography{custom}

\end{document}